\documentclass{article}
\usepackage{ijcai18}

\usepackage{times}
\usepackage{xcolor}
\usepackage{soul}
\usepackage[utf8]{inputenc}
\usepackage[small]{caption}
\usepackage{latexsym}
\usepackage{amsmath}
\usepackage{amssymb}
\usepackage{graphicx}
\usepackage{todonotes}

\usepackage{url}
\usepackage[inline]{enumitem}

\DeclareMathOperator{\Tr}{tr}

\title{
Review Helpfulness Prediction with Embedding-Gated CNN
}

\iftrue
\author{
Cen Chen$^1$, 
Minghui Qiu$^2$, 
Yinfei Yang\thanks{Yinfei Yang is now with Google.}, 
Jun Zhou$^1$,
Jun Huang$^2$,
Xiaolong Li$^1$,
Forrest S. Bao$^3$
\\ 
$^1$ Ant Financial Services Group, Hangzhou, China \\
$^2$ Alibaba Group, Hangzhou, China\\
$^3$ Department of Computer Science, Iowa State University, Ames, IA, USA  \\
\{cecilia.cenchen@gmail.com\}
}
\fi

\begin{document}

\maketitle

\begin{abstract}
Product reviews, in the form of texts dominantly, significantly help consumers finalize their purchasing decisions.
Thus, it is important for e-commerce companies to predict review helpfulness to present and recommend reviews in a more informative manner.
In this work, we introduce a convolutional neural network model that is able to extract abstract features from multi-granularity representations. Inspired by the fact that different words contribute to the meaning of a sentence differently, we propose to learn word-level embedding-gates for all the representations. Furthermore, while some product domains/categories have rich user reviews, other domains do not. To help domains with deficient data, we integrate our model into a cross-domain relationship learning framework for effectively transferring knowledge from other domains.  Extensive experiments show that our model yields better performance than the existing methods.
\end{abstract}

\section{Introduction}
Product reviews, primarily texts, are an important information source for consumers to make purchase decisions. 
Hence, it makes great economical sense to quantify the quality of reviews and present consumers more useful reviews in an informative manner. 
Growing efforts from both academia and industry have been invested on the task of review helpfulness prediction ~\cite{Martin2014,Yang2015,Yang2016,liu-EtAl:2017:EMNLP20174}.

Pioneering work hypothesizes that helpfulness is an underlying property of the text, and uses handcrafted linguistic features to study it. 
For example, \cite{Yang2015} and \cite{Martin2014} examined semantic features like LIWC, INQUIRER, and GALC. Subsequently,  aspect-~\cite{Yang2016} and argument-based~\cite{liu-EtAl:2017:EMNLP20174} features are demonstrated to improve the prediction performance. 

Inspired by the remarkable performance of Convolutional Neural Networks (CNNs) on many tasks in natural language processing, here we employ CNN for review helpfulness prediction task. To better enhance the performance of a vanilla CNN over this task, besides \textit{word-level} representation, we further leverage multi-granularity information, i.e., \textit{character-} and \textit{topic-level} representations. Character-level representations are notably beneficial for alleviating the out-of-vocabulary problem~\cite{ballesteros2015improved,kim2016character,cen2018}, 
while aspect distribution provides another semantic view on the words~\cite{Yang2016}. 

One research question here is whether embeddings shall be treated equally in the CNN. 
Intuitively, different words contribute to the helpfulness of a review in different intensity or importance levels. 
For example, descriptive or semantic words (such as ``great battery life" or ``versatile function") are more informative than general background words like ``phone''. 
Correspondingly, we propose a mechanism called \textit{word-level gating} to weight embeddedings\footnote{
{The gates are applied over all three types of word representations (i.e., character-, word-, and topic-based) for all words.}
}.
Gating mechanisms have been commonly used to control the amount that a unit updates the activation or content in recurrent neural networks~\cite{chung2014empirical}.
Our word-level gates can be automatically learned in our model and help differentiate the important and non-important words. The resulting model is referred to as \textit{Embedding-Gated CNN (EG-CNN)}.

A gating mechanism empowers CNN in two folds. 
First, extensive experiments show that our proposed EG-CNN model greatly outperforms hand-crafted features,  ensemble models of hand-crafted features,  and vanilla CNN models. 
Second, such gating mechanism selectively memorizes the input representations of the words and scores the relevance/importance of such representations, to provide insightful word-level interpretations for the prediction results. The greater a gate weights, the more relevant the corresponding word is to review helpfulness. 

It is common that some product domains/categories have rich user reviews while other domains do not.
For example, the ``Electronics'' domain from Amazon.com Review Dataset~\cite{Mcauley} has more than 354k labeled reviews, while ``Watches'' has less than 10k.
{
Exploiting cross domain relationships to systematically transfer knowledge from \textit{related} domains with sufficient labeled data will benefit the task on domains with limited reviews.
}
It is worth noting that, existing studies on this task only focus on a single product category or largely ignore the inner correlations between different domains.
In previous work, some features are domain-specific while others can be shared. 
For example, image quality features are only useful for cameras~\cite{Yang2016}, while semantic features and argument-based features are applicable to all domains~\cite{Yang2015,liu-EtAl:2017:EMNLP20174}.

While there are some common practices, such as using a shared neural network, to transfer knowledge between domains~\cite{mou:EMNLP2016,yang:iclr2017}, 
  \textit{domain correlations} must be established 
before the knowledge can be transferred properly in our task. Otherwise, transferring the knowledge from a wrong source domain 
may backfire.
We thus provide a holistic solution to both domain correlation learning and knowledge transfer learning by incorporating a domain relationship learning module in our framework.
Experiments show that our final model can correctly tap into domain correlations and facilitate the knowledge transfer between correlated domains to further boost the performance.

The rest of the paper is organized as follows. Section~\ref{sec:model} formally defines the problem and presents our model. Section~\ref{sec:exp} showcases the effectiveness of the proposed model in the experiments. Section~\ref{sec:relwork} presents related work, and finally Section~\ref{sec:conclude} concludes our paper.

\section{Model}\label{sec:model}
We define review helpfulness prediction as a regression task to predict the helpfulness score given a review.
The ground truth of helpfulness is determined using the ``a of b approach'': a of b users think a review is helpful. 

Formally, we consider a cross-domain review helpfulness prediction task where we have a set of labeled reviews from a set of source domains and a target domain. 
We seek to transfer the knowledge from other domains
with rich data to a target domain. 
For a review $\mathbf{X}^k$, our goal is to predict its helpfulness score $y^k$, where $k \in [0..K]$ is the domain label indicating which domain the data instance is from. 



\subsection{Word, Character, and Aspect Representations}
A review $\mathbf{X}$ consists of a sequence of words, i.e., $\mathbf{X} = ({x}_{1}, {x}_{2}, \ldots, {x}_{m})$.
Following the CNN model in~\cite{Kim14f}, for words in a review $\mathbf{X}$,
we first lookup the embeddings of all words $(\textbf{e}_1, \textbf{e}_2, \ldots, \textbf{e}_m)$
from a embedding matrix ${E} \in \mathbb{R}^{|V| \times D}$ where $|V|$ is vocabulary size and $D$ is embedding dimension, and $\mathbf{e}_{i} \in \mathbb{R}^{D\times1}$. This word embedding matrix is then fed into a convolutional neural network to obtain an output representation. This is a typical \textit{word embedding} based model. 

In many applications, such as text classification~\cite{BojanowskiGJM16} and machine reading comprehension~\cite{BiDAF}, it is beneficial to enrich word embeddings with subword information. Inspired by that, we consider to use a character embedding layer to obtain \textit{character embeddings} to enrich word representations. Specifically, the characters of the $i$-th word $x_i$ are embedded into vectors and then fed into another convolutional neural network to obtain a fixed-sized vector $\texttt{CharEmb}(x_i)$. 

A recent work in~\cite{Yang2016} shows that  extracting the aspect/topic distribution from the raw textual contents does help the task of review helpfulness prediction. The reason is that many helpful reviews tend to talk about certain aspects, like `brand', `functionality', or `price', of a product. Inspired by this, we enrich our word representations by aspect distributions. We adopt the model in~\cite{Yang2016} to learn aspect-word distribution $\Phi\in \mathbb{R}^{|A|\times |V|}$, where $|A|$ is aspect size and $|V|$ is the size of vocabulary. 
A word-aspect representation $\Phi'\in \mathbb{R}^{|V| \times |A|}$ is obtained by row-wise normalization of the matrix $\Phi^\top$. Then for each word $x_i$ in input review $X$, we obtain aspect representation by looking up the matrix  $\Phi'$ to get $\Phi'_i \in \mathbb{R}^{|A|\times 1}$.

Formally, for an input review $\mathbf{X}$, we obtain its representation as:
\begin{align}
  \mathbf{e}_X  &=[\mathbf{e}'_1, \mathbf{e}'_2, \ldots, \mathbf{e}'_m], \\  
  \mathbf{e}'_i &= \mathbf{e}_i \oplus \texttt{CharEmb}(x_i) \oplus \Phi'_i, \quad\forall i\in[1..m],
\end{align}
where $\mathbf{e}_i$, $\texttt{CharEmb}(x_i)$, and $\Phi'_i$ represent word-level, character-level, and topic-level representations respectively, and $\oplus$ is a stacking operator.
Note that $m$ ($=100$ in this paper) is the sentence length limit. Sentences shorter than $m$ words will be padded while sentences longer than $m$ words will be truncated.

\subsection{Embedding-gated CNN (EG-CNN)}
Because some words play more important roles in review helpfulness prediction, for example, descriptive or semantic words (such as ``great battery life" or ``versatile function") will be more informative than general background words like `phone'. Hence, we propose to weight the input word embeddings. Specifically we propose a gating mechanism to weight each word in our model. The word-level gate is obtained by feeding the input embeddings to a gating layer, where the gating layer is essentially a fully-connected layer with weight $\mathbf{W}_g$ and bias $b_g$.

Formally, for input $X$, we obtain its representation as follows:
\begin{align}
  \mathbf{e'}_X  &=[g_1 \mathbf{e}'_1, g_2\mathbf{e}'_2,  \ldots, g_m\mathbf{e}'_m],\\
  g_i &= \sigma(\mathbf{W}_g^\top \mathbf{e}'_i + b_g), \quad\forall i\in[1..m], 
\end{align}
where $\sigma$ is a sigmoid activation function.

Next, we stack a 2-D convolutional layers and a 2-D max-pooling layers on the matrix $\mathbf{e}'_X$  to obtain the hidden representation $\mathbf{h}_X$. Multiple filters are used here. For each filter, we obtain a hidden representation:
\begin{eqnarray}
h = \text{CNN} \Big( e_X, \text{filterSize}=\big(f, D, C \big)\Big) \notag
\end{eqnarray}
where $f \in \{2,3,4,5\}$ is window size, $D$ is embedding dimension, $C$ is channel size, and $\text{CNN}(\cdot)$ represents a convolutional layer followed by a max-pooling layer. All the representations will then be concatenated to form the final representation $\mathbf{h}_X$. We refer our base model as Embedding-Gated CNN (EG-CNN), where EG-CNN learns a hidden feature representation $\mathbf{h}_X$ for an input $X$.



\subsection{Cross-Domain Relationship Learning}
If we treat all the domains as the same domain, we can build an unified model for our task. Specifically, our target here is to optimize the following objective:
\begin{eqnarray}
 l = \sum_k \sum_{X} (\mathbf{U}^\top \texttt{EG-CNN}(\mathbf{X}^k) - y^k)^2 + l_{reg},
 \label{eq:no_cross_domain_relationship}
\end{eqnarray}
where $\mathbf{U}$ is the output layer, $\mathbf{X}^k$ is the input from domain $k$, $y^k$ is the corresponding label, $l_{reg}$ is a regularization term. 

The formulation in Eqn.~(\ref{eq:no_cross_domain_relationship}) is limited because it does not take the difference of domains into consideration. To utilize the multi-domain knowledge, we convert the method above to a multi-domain setting where we assume an output layer $\mathbf{W}_k$ for each domain $k$. 
While still a unified model to learn universal feature representation, our new approach has two output layers $\mathbf{U}$ and $\mathbf{W}$ to model domain \textit{commonalities} and \textit{differences} respectively.

Furthermore, we explicitly model a domain correlation matrix $\mathbf{\Omega} \in \mathbb{R}^{K\times K}$, where $\mathbf{\Omega}_{i,j}$ is the correlation  between domains $i$ and $j$. Following the matrix-variate distribution setting from~\cite{zhang:uai2010}, our objective is to optimize the trace of the matrix product $\text{tr} (\mathbf{W\Omega}^{-1}\mathbf{W}^\top)$. This shows, when domain $i$ and domain $j$ are close, i.e., $\mathbf{W}_i$ is close to $\mathbf{W}_j$,
the model tends to learn a large $\mathbf{\Omega}_{i,j}$ in order to minimize the trace. In all, our objective is defined as follows:
\begin{align}
 l \quad= &\sum_k \sum_{\mathbf{X}} ((\mathbf{U}+\mathbf{W}_k)^\top \texttt{EG-CNN}(\mathbf{X}^k) - y^k)^2 +  \nonumber \\
     & \lambda_1 \text{tr} (\mathbf{W\Omega}^{-1}\mathbf{W}^\top) + \lambda_2 l_{reg}, \nonumber \\
s.t. \quad& \mathbf{\Omega} \succeq \mathbf{0}, \quad \Tr(\mathbf{\Omega}) = 1.
\label{eqn:objective}
\end{align}
where $\Tr(\cdot)$ gets the trace of a  matrix, $l_{reg}$ is a regularization term, $\lambda_1$ and $\lambda_2$ are weight coefficients.

\begin{figure}[!htbp]
	\centering
	\includegraphics[width=1\columnwidth]{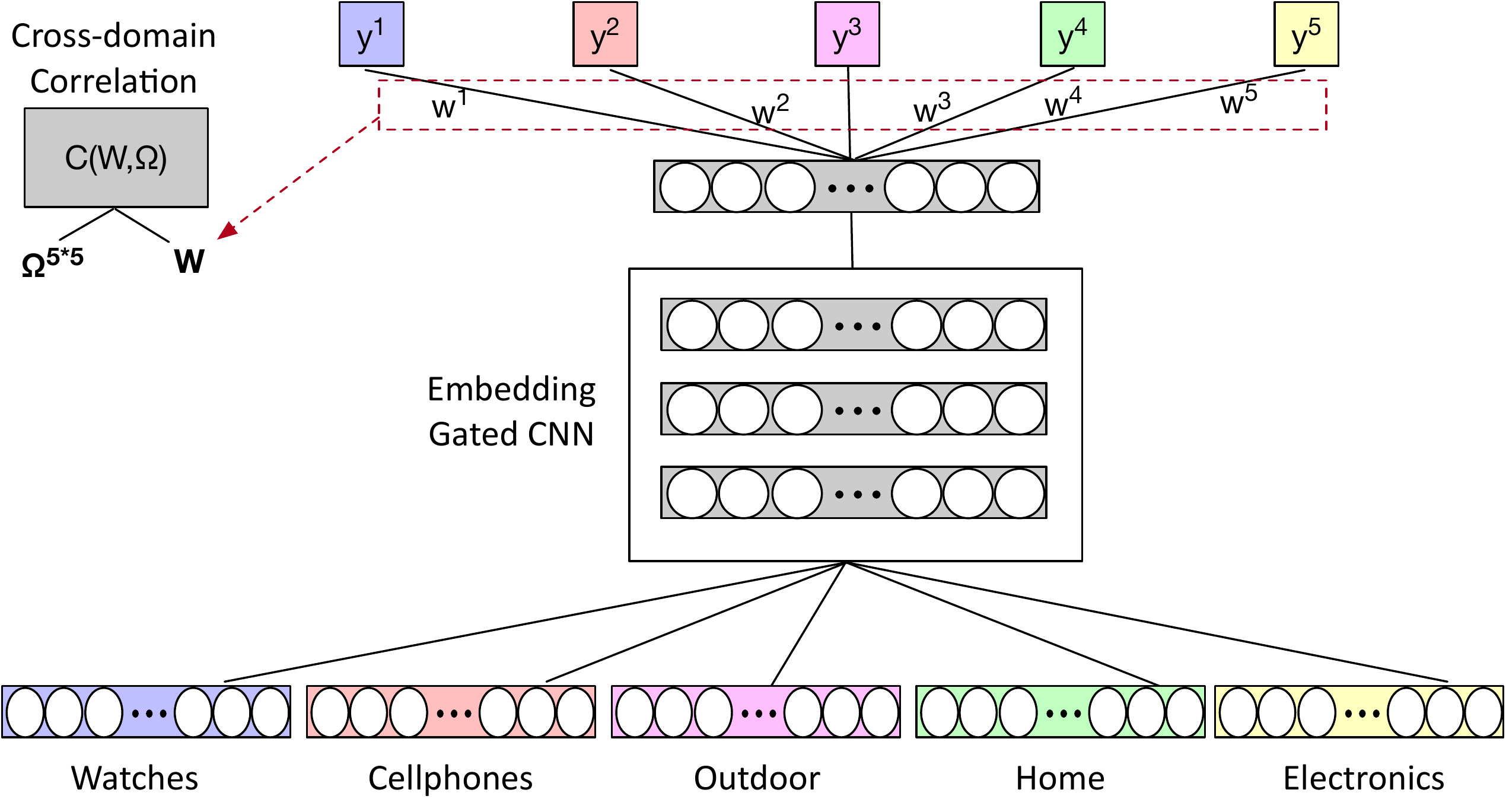}
	\caption{Our final model with cross-domain relationship learning.}
	\label{fig:final_model}
\end{figure}

Our final model is presented in Figure~\ref{fig:final_model}, where we use EG-CNN as our base model, and further consider cross-domain correlation and multiple domain training. Note that, if we set $\mathbf{\Omega}$ as an identity matrix (no domain correlation) and $\mathbf{U}=\mathbf{0}$ (no shared output layer), the multi-domain setting is degenerated to a fully-shared setting in \cite{mou:EMNLP2016}. The limitation of the fully-shared setting is that it ignores domain relationships. However, in practise, we may think ``Electronics" is helpful to ``Home'' and ``Cellphones'' domains, but may not be so helpful as for ``Watches" domain. With our model, we seek to automatically capture such domain relationships and use that information to help boost model performance.

\section{Experiments}\label{sec:exp}

Reviews from 5 different categories in the public Amazon Review Dataset~\cite{Mcauley} 
 are used in the experiments.
Data statistics are summarized in Table~\ref{tab:dataset}.
\begin{table}[!htbp]
\centering
\caption{Amazon reviews from 5 different categories.}
\small
\begin{tabular}{l|r r }
\hline
\parbox{2.5cm}{General category} & \parbox{1.9cm}{\# of reviews$\quad$ ($>$ 5 votes)} & \parbox{1.9cm}{\# of reviews}\\
\hline
Watches       & 9,737   &    68,356 \\
Cellphones (Phones)   & 18,542  &    78,930 \\
Outdoor       & 72,796   &  510,991 \\
Home          & 219,310  &    991,784 \\
Electronics (Elec.)    & 354,301 & 1,241,778\\
\hline
\end{tabular}
\label{tab:dataset}
\end{table}

In our model, we initialize the lookup table $\mathbf{E}$ with pre-train word embeddings from \textit{GloVe}~\cite{pennington:emnlp2014} with $D=100$.
For aspect representations, we adopt the settings from~\cite{Yang2016} to set the topic size to 100.
For EG-CNN, the activation function is ReLU,  the channel size is set to 128, and
AdaGrad~\cite{duchi:jmlr2011} is used in training with an initial learning rate of 0.08.

Following the previous work, all experiment results are evaluated using {correlation coefficients} between the predicted helpfulness score and the ground truth score. The ground truth scores are computed by ``a of b approach'' from the dataset, indicating the percentage of consumers thinking a review as useful.

\subsection{Comparison with Linguistic Feature Baselines and CNN Models}
Our proposed EG-CNN model is compared with the following baselines:
\begin{itemize}
    \item STR/UGR/LIWC/INQ/ASP: Five regression baselines that use handcrafted features such as ``STR",  ``UGR",  ``LIWC", ``INQ"~\cite{Yang2015} and aspect-based features ``ASP"~\cite{Yang2016}", respectively;
    \item Fusion$_1$: ensemble model with ``STR", ``UGR", ``LIWC", and ``INQ" features~\cite{Yang2015}
    \item Fusion$_2$ : Fusion$_1$ with additional ``ASP" features~\cite{Yang2016}. 
    \item CNN: the vanilla CNN model~\cite{Kim14f} 
    with word-level embedding only;
    \item CNN$_{c}$: the vanilla CNN model with character-based representation ~\cite{cen2018};
    \item CNN$_{ca}$: the vanilla CNN model with character- and topic- based representations.
    \item EG-CNN: our final model with word-level, character-level, and topic-level representations in a gating mechanism.
\end{itemize}

\begin{table}[!h]
\small
\centering
\caption{Comparison with linguistic feature baselines and CNN models.}
\begin{tabular}{l|r r r r r}
\hline
&    Watches    &    Phone    &    Outdoor    &    Home    &    Elec.    \\
\hline \hline
STR    &    0.276     &    0.349     &    0.277     &    0.222     &    0.338     \\
UGR    &    0.425     &    0.466     &    0.412     &    0.309     &    0.355     \\
LIWC    &    0.378     &    0.464     &    0.382     &    0.331     &    0.400     \\
INQ.    &    0.403     &    0.506     &    0.419     &    0.366     &    0.405     \\
ASP    &    0.406     &    0.437     &    0.385     &    0.283     &    0.406     \\\hline
Fusion$_1$    &    0.488     &    0.539     &    0.497     &    0.432     &    0.484     \\
Fusion$_2$        &    0.493     &    0.550     &    0.501     &    0.436     &    0.491     \\\hline
CNN    &    0.480     &    0.562     &    0.501     &    0.459     &    0.524     \\
CNN$_c$    &   0.495     &    0.566     &    0.511    &    0.464     &    0.521     \\
CNN$_{ca}$   &   0.497     &    0.567     &    0.524    &    0.476     &    0.537     \\
EG-CNN    &    \textbf{0.515}     &    \textbf{0.585}     &    \textbf{0.555}     &    \textbf{0.541}     &    \textbf{0.544}     \\
\hline
\end{tabular}
\label{tab:exp1}
\end{table}

Table~\ref{tab:exp1} shows several interesting observations that validate our motives behind this work. First, all the CNN based models consistently outperform non-CNN models, indicating their expressiveness over handcrafted features. Second, CNN$_{c}$ outperforms CNN when data is relatively insufficient (e.g., the domains ``Watches'' and  ``Phones'') and loses its edge on domains of abundant data (e.g., the domain ``Electronics''). 
This is because when data size is smaller, 
the out-of-vocabulary problem (OOV) is more severe, and character-based representation is more beneficial. Third, CNN$_{ca}$ consistently outperforms CNN$_{c}$, showing that adding topic-based representations can further help the task. Last but not least, our proposed EG-CNN outperforms all CNN variants, which justifies the necessity of adding embedding gates. 
This further supports the importance of considering embedding gates.
In all cases, EG-CNN significantly outperforms the baselines and yields better results than all CNN variants.

\subsection{Comparison with Cross-domain Models}
To evaluate the effectiveness of our domain relationship learning, we compare our proposed full model against the following two baselines: the target-only model that uses only data from the target domain, and the fully-shared model that uses a fully shared neural network~\cite{mou:EMNLP2016} for all domains. 
{
All three models use EG-CNN as the base model.
}
 \begin{table}[!h]
     \small
     \centering
    \caption{
    {
    Comparison with target-only and fully-shared models. 
    Note that both fully-shared and our model use the multi-task setting, where all domains are jointly modeled and learned. 
    }
    }
    \begin{tabular}{l|r r r r r r}
        \hline
        & Watches & Phones & Outdoor & Home & Elec.   \\
        \hline\hline
        Target-only & 0.515 & 0.585 & 0.555 & 0.541 & 0.544 \\
        Fully-shared &  0.522 & 0.580 & 0.551 & 0.518 & 0.534 \\
        Ours    & \textbf{0.535} & \textbf{0.592} &    \textbf{0.561} & \textbf{0.544} & \textbf{0.548} \\
        \hline
    \end{tabular}

    \label{tab:exp3}
\end{table} 


In all experiments, our model consistently achieves better results than both target-only and fully-shared models, supporting the effectiveness and benefit of cross-domain relationship learning. 
The improvement is greater on domains with fewer labeled data, e.g., the ``Watches'' domain. 
The ``Watches'' domain has the least number of reviews and our model shows the most improvement there. 

Interestingly, the fully-shared model performs much worse than the target-only model in the ``Home" domain.
This might be justified by the potential domain shift, under which the fully-shared model may not perform better than the target-only model. 
Because some domains are related while some others are  not, incorporating data from those less related can hardly help, especially when the target domain (such as ``Home") has sufficient data for the target-only model to perform well enough. 
\section{Related Work}\label{sec:relwork}

Recent studies on review helpfulness prediction extract handcrafted features from the review texts. For example, \cite{Yang2015} and \cite{Martin2014} examined semantic features like LIWC, INQUIRER, and GALC. Subsequently, aspect-~\cite{Yang2016} and argument-based~\cite{liu-EtAl:2017:EMNLP20174} features are demonstrated to improve the prediction performance. 
These methods require prior knowledge and human effort in feature engineering and may not be robust for new domains. 
In this work, we employ CNNs~\cite{Kim14f,zhang2015character} for the task, which is able to automatically extract deep features from raw text content. 
As character-level representations are notably beneficial for alleviating the out-of-vocabulary~\cite{ballesteros2015improved,kim2016character}, while aspect distribution provides another semantic view on words~\cite{Yang2016}, we further enrich the word representation of CNN by adding multi-granularity information, i.e., character- and aspect-based representations.
As different words may play different importance on the task, we  consider to weight word representations by adding word-level gates. 
Gating mechanisms have been commonly used in recurrent neural networks to control the amount of unit update the activation or content and have demonstrated to be effective~\cite{chung2014empirical,dhingra2016gated}.
Our word-level gates help differentiate important and non-important words. The resulting model, referred to as embedding-gated CNN, has shown to significantly outperform the existing models.  

It is common that some domains have rich user reviews while other domains may not. To help domains with limited data, we study cross-domain learning (transfer learning~\cite{pan:tkde2010} or multi-task learning~\cite{zhang2017survey}) for this task. Transfer learning and multi-task learning have been extensively studied in the last decade. With the popularity of deep learning, a great amount of Neural Network (NN) based methods are proposed for TL~\cite{yosinski:nips2014}. A typical framework is to use a shared NN to learn shared features for both source and target domains~\cite{mou:EMNLP2016,yang:iclr2017}. Another approach is to use both a shared NN and domain-specific NNs to derive shared and domain-specific features~\cite{liu:acl2017}. 
A multi-task relationship learning method is introduced in~\cite{zhang:uai2010}, which is able to uncover the relationship between domains. Inspired by this, we adopt the relationship learning module to our EG-CNN framework to help model the correlation between different domains. 

To the best of our knowledge, our work is the first to propose gating mechanism in CNN and to study cross-domain relationship learning for review helpfulness prediction. 


\section{Conclusion}\label{sec:conclude}
In this work, we tackle review helpfulness prediction using two new techniques, i.e., embedding-gated CNN and cross-domain relationship learning. We built our base model on CNN with word-, character- and topic-based representations.
On top of this model, 
domain relationships were learned to better transfer knowledge across domains. 
The experiments showed that our model significantly outperforms the state of the art. 


\bibliographystyle{named}
\bibliography{ijcai18}

\end{document}